\documentclass[10pt, a4paper]{article}

\usepackage{lrec-coling2024} 

\usepackage{enumitem}
\setlist[itemize,1]{left=0.2em, label=$\bullet$, itemsep=6pt}

\usepackage{graphicx,subfig}
\usepackage{float}

\usepackage{hyperref}

\newcommand{\vago}[0]{\texttt{VAGO}}
\newcommand{\vagon}[0]{\texttt{VAGO-N}}
\newcommand{\mBERT}[0]{\texttt{mBERT}}
\newcommand{\BERT}[0]{\texttt{BERT}}
\newcommand{\camemBERT}[0]{\texttt{CamemBERT}}
\newcommand{\gatecloud}[0]{\texttt{GATE Cloud}}
\newcommand{\gate}[0]{\texttt{GATE}}
\newcommand{\obsinfox}[0]{\texttt{OBSINFOX}}
\newcommand{\pe}[1]{\textcolor{black}{#1}}

\title{A Multi-Label Dataset of French Fake News:\\
Human and Machine Insights}

\name{Benjamin Icard$^{1,2}$, {\bf \large François Maine$^{1,3}$}, {\bf \large Morgane Casanova$^{4}$}, {\bf \large Géraud Faye$^{5,6}$} \\ {\bf \large Julien Chanson$^{7}$}, {\bf \large Guillaume Gadek$^{5}$}, {\bf \large Ghislain Atemezing$^{8}$} \\ {\bf \large François Bancilhon$^{9}$}, {\bf \large and Paul \'Egr\'e$^{2}$}}

\address{\vspace{0.1in} \\
          $^{1}$LIP6, Sorbonne Université, CNRS, France - $^{2}$Institut Jean-Nicod, CNRS, ENS-PSL, EHESS, France 
          \\
          $^{3}$Freedom Partners, France -  
          $^{4}$Université de Rennes, CNRS, Inria, IRISA, France 
          \\
          $^{5}$Airbus Defence and Space, France -       $^{6}$Universit\'e Paris-Saclay, CentraleSup\'elec, MICS, France
          \\
          $^{7}$Mondeca, France - 
          $^{8}$European Union Agency for Railways, France
          \\
          $^{9}$Observatoire des Médias, France
  \vspace{0.3in}        
}

\abstract{We present a corpus of 100 documents, \textcolor{black}{named \obsinfox}, selected from 17 sources of French press considered unreliable by expert agencies, annotated using 11 labels by 8 annotators. By collecting more labels than usual, by more annotators than is typically done, we can identify features that humans consider as characteristic of fake news, and compare them to the predictions of automated classifiers. We present a topic and genre analysis using \gatecloud, indicative of the prevalence of satire-like text in the corpus. We then use the subjectivity analyzer \vago, and a neural version of it, to clarify the link between ascriptions of the label Subjective and ascriptions of the label Fake News.
The annotated dataset is available online at the following url: \textcolor{black}{\url{https://github.com/obs-info/obsinfox}}
 \\ \newline }

\begin{document}

 \maketitleabstract

\section{Introduction}

One of the challenges raised by fake news is that the very notion of fake news is multidimensional. It includes fabrication, satire, but also mistaken reports, and often just biased or partisan information \cite{tandoc2018defining, gelfert2018fake, zhou2018fake}.

Notwithstanding that complexity, algorithms trained to detect fake news typically rely on datasets involving just two labels, such as ``biased'' vs ``legitimate'' (viz. ISOT\footnote{\url{https://onlineacademiccommunity.uvic.ca/isot/\#datasets}}), with no indication of the type of fake news in question, let alone the cues used to explain the labels. In order to get reliable detectors of fake news, however, it matters to use datasets and labels that are sufficiently precise in order to inform classifiers along several dimensions. \pe{Some multi-label fake news datasets are available, such as LIAR \cite{wang2017liar} (6 labels), or the Brazilian dataset of \cite{morais2019deciding} (4 labels). In the former, labels qualify levels of truth (following the \url{politifact.com} guidelines) and in the in the latter, the \textit{legitimate-biased} distinction is crossed with the presence or absence of satire. However, except for that feature, these labels do not pertain to stylistic information}.

In this paper, we report on the constitution and annotation of a corpus of French press \textcolor{black}{named \obsinfox}, selected from websites categorized by expert organizations as unreliable, and so as good candidates to include biased, exaggerated, or even factually false statements. While \textcolor{black}{\obsinfox\ } is limited in size (100 documents), 
our goal was to obtain a rich dataset, by considering 11 labels for annotation, and then by asking 8 annotators to annotate it.

The aim was twofold: on the one hand, we intend to identify which labels are most informative of the status of a text. On  the other, we are interested in finding the cues in those texts that best explain their classification by humans and then by machines as containing fake news or not.







Section \ref{sec:materials} explains the selection of the corpus \textcolor{black}{\obsinfox}, the choice of the labels, and the guidelines and method for collection of the annotations. Section \ref{sec:topic} gives an analysis of the topics and genres of the corpus using \gatecloud, and Section \ref{sec:annotations} presents an analysis of the human annotations and their relations. In Section \ref{sec:vago}, finally, we examine the way in which the label ``Fake News'' is ascribed in relation to other labels, in particular to ``Subjective'', ``False'' and ``Exaggerated''. 
Toward that goal, we use the text analyzer \vago\ to relate scores of linguistic subjectivity with human scores on the label ``subjective''. We validate this approach by using a neural version of \vago, trained on a distinct corpus ``FreSaDa'' \cite{IonescuChifu2021IJCNN}, of satirical news.




\section{Corpus and Labels}\label{sec:materials}

\begin{figure}[h]
\begin{footnotesize}
\fbox{
\parbox{7.3cm}{
\begin{itemize}
\item[$\bullet$] \textcolor{black}{\textbf{Fake News:}} the article describes at least a false or exaggerated fact.

\item[$\bullet$] \textbf{Places, Dates, People:} the article mentions at least one place, date or person.

\item[{$\bullet$}] \textcolor{black}{\textbf{Facts:}} the article reports at least one fact, i.e. a state of affairs or event, which may be true or false. 

\item[{$\bullet$}] \textcolor{black}{\textbf{Opinions:}} the article expresses at least one opinion.

\item[{$\bullet$}] \textcolor{black}{\textbf{Subjective:}} the article contains more opinions than facts.

\item[$\bullet$] \textbf{Reported Information:} the information of the article is reported by another person or source, and is not directly endorsed. 

\item[$\bullet$] \textbf{Sources Cited:} the article cites at least one source, for at least one fact.

\item[{$\bullet$}] \textcolor{black}{\textbf{False Information:}} the article contains at least one false fact. 

\item[$\bullet$] \textbf{Insinuation:} the article suggests a certain reading of a fact, without saying so explicitly.

\item[$\bullet$] \textbf{Exaggeration:} the article  describes a real fact with exaggeration.

\item[$\bullet$] \textbf{Offbeat Title:} the article has a misleading headline not accurately reflecting the content of the article.

\end{itemize}}}
\end{footnotesize}
\caption{Description of the 11 labels selected for the annotation task.}
\label{fig:annot-labels}
\end{figure}



The dataset \obsinfox\ was compiled from online sources of French press presented as unreliable and prone to propagating fake news by NewsGuard\footnote{\url{https://www.newsguardtech.com/}} and Conspiracy Watch\footnote{\url{https://www.conspiracywatch.info/}} in particular. The time period covered goes from 2010 to 2023, but is mostly focused on the 3 last years. Exactly 100 articles were selected for the study. 
That sample originated in a larger corpus of 54,845 online articles, itself the result of keeping only the 17 most popular French sources among the 40 involved in an original corpus of 101,200 articles. A pilot study involved the selection of 906 articles within the 54,845 articles in order to conduct a first human annotation task with 4 annotators and 26 labels. We used the \texttt{TfidfVectorizer} transformer to pre-select 120 articles among the 54,845 articles, half of which with a probability of reporting fake news above .8 according to the predictor, half with a probability below .2. Among those 120, 100 were retained after elimination of 20 articles too short or uninterpretable, with 49 predicted to be fake news, and 51 not. A detailed list of the press sources included can be found in the README file available on the \obsinfox\ repository.



For the labels, Figure \ref{fig:annot-labels} presents them in the order in which annotators had to mark them, with a summary of their definition. \textcolor{black}{The selection of the labels was based on prior meetings, during which the annotators iteratively discussed the procedure and the annotation manual including the definitions provided in Figure \ref{fig:annot-labels}. The annotators eventually agreed on 11 labels after discussing a broader set of 26 labels coming from the pilot study on 906 articles mentioned in section \ref{sec:materials}. 
Another decision was to allow only binary responses instead of more answer types (such as “I don’t know”), but participants were authorized to leave personal comments (eventually removed from the dataset).}

The labels distinguish ``Fake News'' from ``False Information", and define the former more widely (as involving falsity or exaggeration), to distinguish plain falsities  (``Obama was not born in the USA'') from cases of exaggeration involving partially true facts (``inflation skyrockets everywhere"). 
The label ``Offbeat Title" is linked to clickbait detection, usually a marker of exaggeration or distortion. The label ``Opinion'' looks for the occurrence of at least one opinion sentence, ``Subjective'' concerns whether opinions are prevalent over objective reports. The labels ``Places, Dates, People" and ``Facts" are used to assess whether an article reports at least a factual piece of information or only the author’s opinion. The presence of location, temporal, or nominal information allows fact-checking systems to process the article, in contrast to articles containing only opinions \cite{guo2022survey}. ``Reported Information'' and ``Sources Cited'' are related and can help identify if the information is directly endorsed by the writer and if it comes from secondary sources. Beside ``Exaggeration'', the label ``Insinuation'' was also included to detect indirect derogatory techniques (such as dog-whistle).

The 8 annotators included the designers of the experiment (7 male, age range from 29 to 76).
Only one of them had seen the texts prior to annotating, in order to upload them on the platform. Annotators didn't have access to the url to avoid bias by source, unlike in other datasets (ISOT or \citealt{horne2017just}). The resulting dataset does not present aggregate data (as do PolitiFact and GossipCop, \citealt{shu2018fakenewsnet}) but includes individual annotations grouped by annotator, to give access to individual variability and allow for more refined analyses. 





\section{Topic and Genre Analysis}\label{sec:topic}

\begin{figure}[h]
    \centering
    \includegraphics[width=1.0\linewidth]{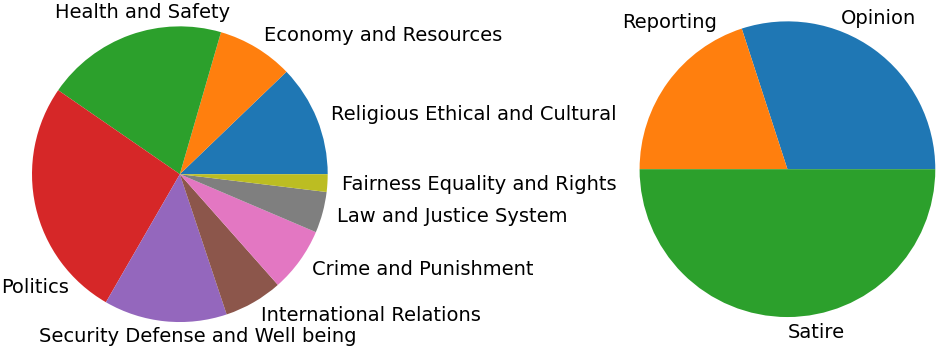}
    \caption{Topic and genre distribution in the corpus.}
    \label{fig:genres_topic}
\end{figure}

Analyses of the corpus were conducted after selection, using \textcolor{black}{pretrained} tools made available by \gatecloud.\footnote{\url{https://cloud.gate.ac.uk/}} More precisely, the topics and genres of articles were detected using an ensemble of \mBERT\ models \cite{wu2023sheffieldveraai}. \textcolor{black}{These tools have been chosen because of their accessibility and the good performance they achieved during the SemEval 2023 Task 3.} The distributions of topics is shown in Figure~\ref{fig:genres_topic}, left. Topics are diverse, but nearly half of the articles deal with Politics and with Health and Safety, followed by Security Defense and Well-being, and Religious, Ethical and Cultural topics. 

\begin{figure}[h]
    \centering
    \includegraphics[width=1.0\linewidth]{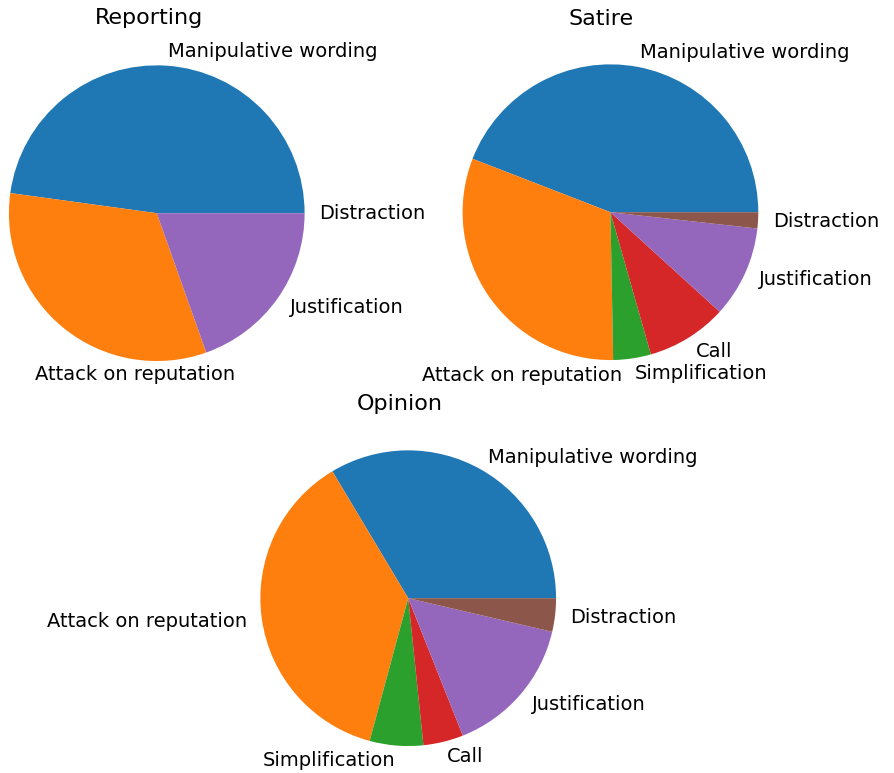}
    \caption{Persuasion techniques by genre.}
    \label{fig:persuasion}
\end{figure}

\textcolor{black}{To further our analysis, we categorized the different articles into genres, following the three-way news categorization proposed by \cite{piskorski2023semeval} into \textit{opinion} pieces, pieces aiming at objective news \textit{reporting}, and \textit{satire} pieces, also using the tools provided by \gate\ (Figure \ref{fig:genres_topic}, right). Half of the chosen articles are written in a satire-like style (only stylistically, as no real satire involving humor is present in the corpus). This confirms the observations made in \cite{horne2017just} about the prevalence of caricature and exaggeration in fake news. Within each type, we looked at the manipulative persuasion techniques inventoried in \cite{piskorski2023semeval}, based on the taxonomy proposed by \cite{da-san-martino-etal-2020-semeval} for propaganda. They include 23 techniques in total, falling into 6 main groups, including so-called manipulative wording, distraction, attack on reputation, call (to act or think), simplification, and [partisan or biased] justification (see Figure~\ref{fig:persuasion}).}

The distributions of persuasion techniques are approximately the same for opinion articles and for satire-like articles. They differ by the number of persuasion techniques used by articles, with opinion and satire-like containing respectively a mean of 3.4 and 4.6 persuasions techniques by article. In the reporting articles, less diverse persuasion techniques are found, which is to be expected as they are more factual. However, the number of persuasion techniques found is relatively high (2.3 per article) for a factual content, even if it is lower than for opinion pieces.


\section{Human Annotations}\label{sec:annotations}






In order to assess the quality of annotations, we measured the inter-annotator agreement among the 8 human annotators, using two  distinct measures. First we calculated Fleiss's kappa for each document, in order to shed light on the overall reliability of their collective judgments. For the 11 annotation labels given in Figure \ref{fig:annot-labels}, we obtained a mean value per document of $\kappa=0.4659074$, showing moderate agreement between annotators of the panel. 

\begin{figure}[htb]
\subfloat[\centering Mean inter-annotator scores per label.]{\label{fig:all}\includegraphics[width=0.47\textwidth]{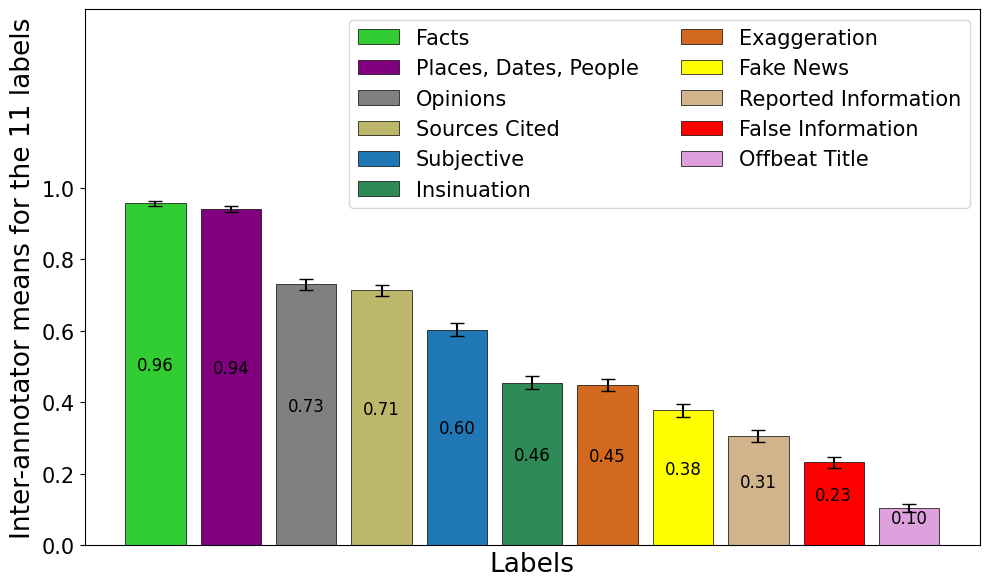}}%
\qquad

\subfloat[\centering Mean inter-annotator agreement per label.]{\label{fig:agreement}\includegraphics[width=0.47\textwidth]{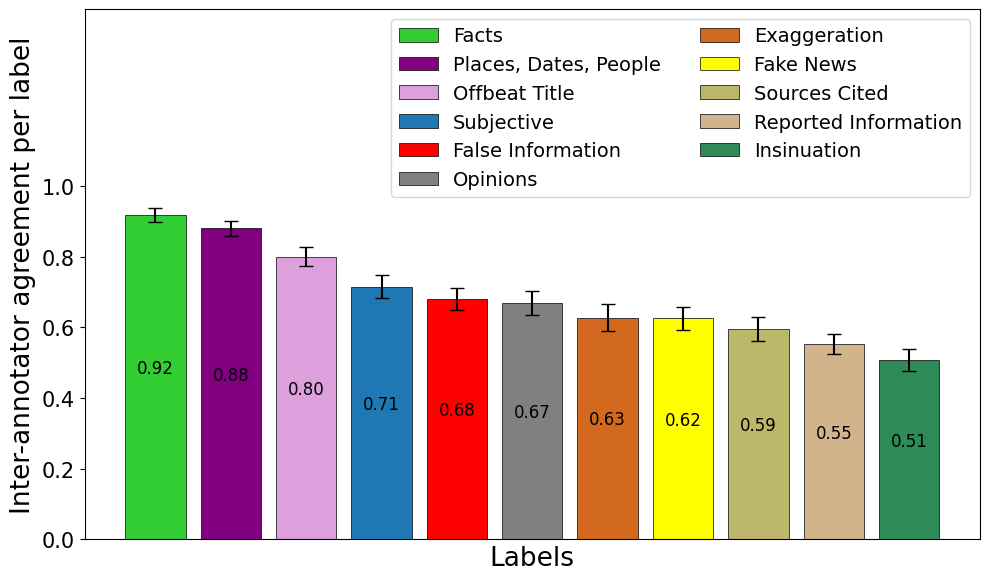}}%

\caption{\label{fig:comaprison} Mean scores and agreement by label (error bars=standard error of the mean).}
\end{figure}

The second method we used, displayed in Figure \ref{fig:agreement}, consisted of rescaling the percentage of agreement between annotators: for each document, we computed the proportion $x$ of answers equal to 1, rescaled by the function returning the value $\alpha=|2x-1|$. The rescaling implies that when only half of the annotators agree on a label, the level of agreement is 0. When 75\% go in the same direction, agreement is 0.5, and a value of .7 or above indicates 85\% of agreement or more. \textcolor{black}{Compared to Fleiss’s kappa, the rescaling method is easier to interpret and sheds light on the inter-annotator agreement per article for each of the 11 labels. In addition, Pearson's calculation shows that both metrics are very well correlated ($r=0.94$, $p=3.06e-46$). 
}

As Figure \ref{fig:agreement} shows, all labels reached a mean value above .5. The label ``Facts'' shows the highest agreement, and the label ``Insinuation" the lowest. Other labels of particular interest for us, concretely ``Fake news'', ``False Information'', ``Opinions'', and ``Subjective'', all reach a mean score above .6, with the highest level for ``Subjective'' (0.715).

Figure \ref{fig:all} shows how much on average a label was used across the 100 articles by each annotator (mean of means). The label ``Fake News'' reaches a mean of .38, hence below the proportion predicted at selection, and below the percentage of Satire found by \gatecloud.\footnote{However, a comparison between the labels of the predictor used for selection and majorities on the label ``Fake News'' shows low accuracy of 0.40, 0.40 and 0.36 respectively for levels of agreement $\alpha=.5, =0.75$ and $=1$.} ``False Information'', with a mean of .23, is ascribed less than ``Fake News'', consistently with their definition. 

\begin{figure}[htb]
\includegraphics[scale=.37]
{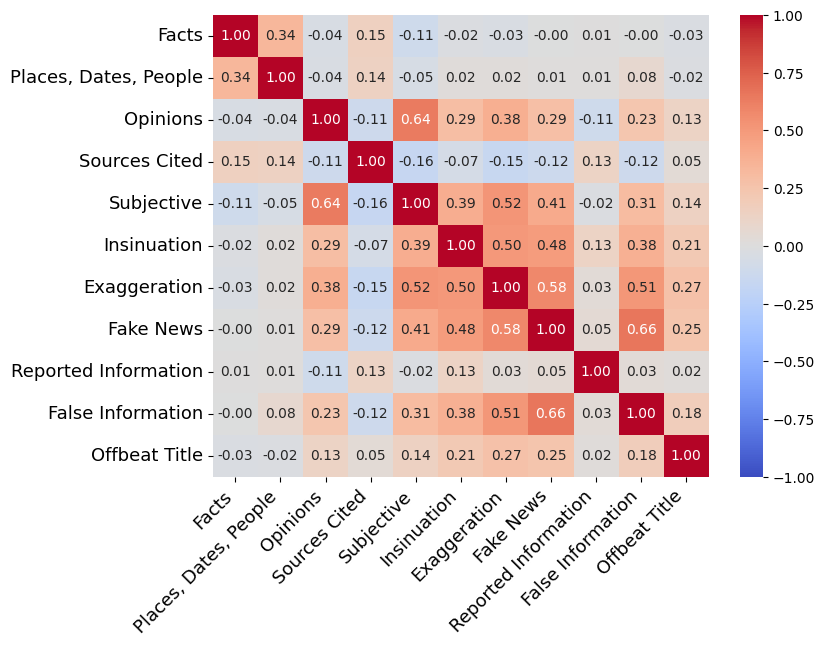}%

\includegraphics[scale=.37]{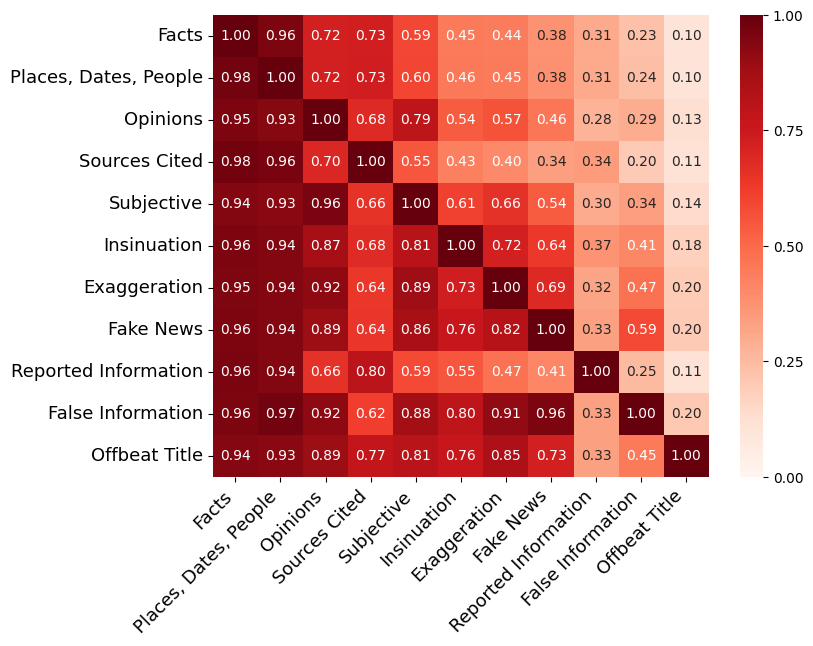}%
\caption{Correlation matrix between the 11 labels (top), and percentage of a row label satisfying a column label (bottom).}\label{fig:corrmatrix}
\end{figure}

Figure \ref{fig:corrmatrix} (top) displays the correlation between the 11 labels and shows that the labels that correlate the most are ``Subjective'', ``Opinion'', ``Insinuation'', ``Exaggeration'', ``Fake News'', and ``False Information''. Figure \ref{fig:corrmatrix} (bottom) also reports, from the 800 judgment profiles, the proportion of a row label $A$ that is (asymmetrically) associated with a column label $B$. 59\% of items tagged as ``Fake news'' are tagged as ``False Information'', versus 96\% of ``False Information'' tagged as ``Fake News''. The proportions of ``Exaggeration''/``Fake News''/``False Information'' tagged as ``Subjective'' are 89\%, 86\%, 88\%. Conversely, the proportion of ``Subjective'' documents tagged as ``Exaggeration''/``Fake News''/``False Information'' is 66\%, 54\%, 34\%. This indicates that while the inference from ``Fake'' or even ``False'' to ``Subjective'' is strong, the converse inference from ``Subjective'' to ``Fake'' and ``False'' is weaker. 



\section{Ascriptions of ``Fake News'' 
}\label{sec:vago}


In light of the associations between the labels ``Fake News'', ``Exaggeration'', and ``Subjective'', and to understand the linguistic cues picked by annotators, we used an automated detector of subjectivity in texts, the \vago\ tool, applied on larger corpora to relate the occurrence of subjective lexicon in text with the detection of fake news \citep{Guelorget2021combining,icard&alTALN2023}. For a given text, \vago\ computes three scores, a score of vagueness, a score of opinion, and a score of relative detail compared to vagueness. \vago\ does not incorporate any world-knowledge, but checks for the occurrence of markers of subjectivity (including first-person pronouns, exclamation marks, and terms of exaggeration or slurs among evaluative adjectives), as well as markers of objectivity (named entities in particular).

\vago\ produces a score of linguistic subjectivity that previous studies have found positively correlated with the label ``biased'' in news articles \cite{Guelorget2021combining,icard&alWI-IAT2023}, and a score of detail-vs-vagueness that previous studies have found negatively correlated with the label ``satirical'' \cite{icard&alWI-IAT2023}. Hence, we hypothesized that larger \vago\ scores of opinion should predict higher use of the labels ``Subjective'', and ``Opinions''. 
For ``Fake News'', however, we expect a weaker association, since falsity is a separate component of that label as defined in the annotation guide.

\begin{figure}[t]
\includegraphics[width=0.24\textwidth]{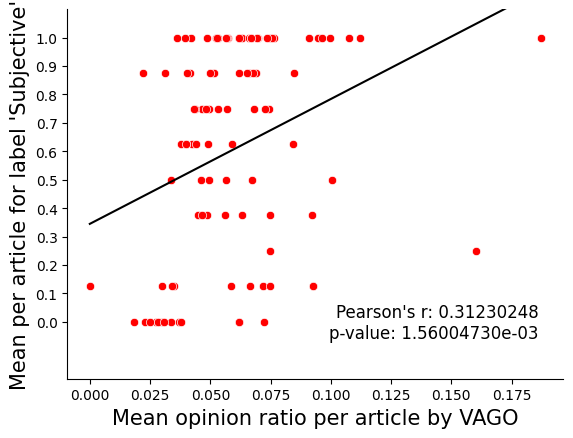}%
\includegraphics[width=0.24\textwidth]{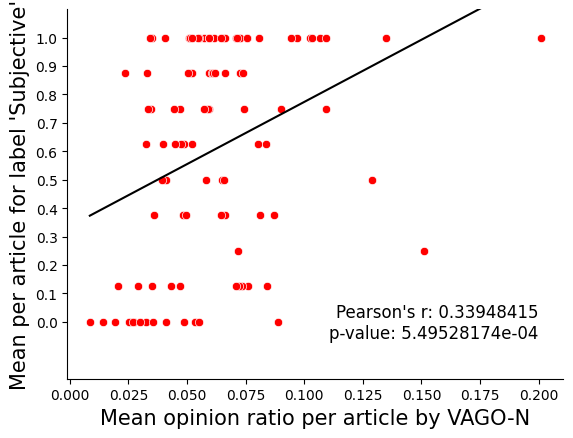}%
\caption{Pearson correlations between the mean opinion score per article provided by \texttt{VAGO} and \texttt{VAGO-N} and the mean inter-annotator score for the label ``Subjective''.}\label{fig:vagons}
\end{figure}

To test those hypotheses, we calculated the correlation between the \vago\ scores for each document in the corpus and the mean inter-annotator scores for the labels ``Subjective'', ``Opinions'', ``Exaggeration'', ``Fake News'', and finally, ``False Information''. We used two sets of \vago\ scores: those produced by the expert system \vago, and the scores produced by a neural clone \vagon. 
This neural version \texttt{VAGO-N} combines the ``\camemBERT-base'' French version \cite{martin2019camembert} of the \BERT\ model \cite{devlin_bert:_2018} (\textit{Batch Size}=5, \textit{Learning Rate}=1e-05, \textit{Epochs}=5) with 3 regression layers and 3 MSE loss functions to predict the scores of vagueness, opinion and detail of sentences. Building the model consisted of using the \vago\ scores on the 141,137 sentences of the French corpus ``FreSaDa''\footnote{\url{https://github.com/adrianchifu/FreSaDa}} \cite{IonescuChifu2021IJCNN}, making the following random selection: 99,022 sentences for training, 21,219 sentences for validation and 21,219 sentences for test. We obtained high performance for the three scores as indicated by the Root Mean Square Error (RMSE) measures: 0.026 for vagueness, 0.028 for opinion, and 0.083 for detail. We obtained similar performances by comparing \vago\ with \vagon\ on the 100 articles (see Table \ref{tab:rmsecomp}), and generally, found close correlation scores between the two versions of \vago\ across the labels tested (Figure \ref{fig:vagons}).

\begin{table}[h]
\begin{center}
{\small
\begin{tabular}[t]{l|c|c}
                  RMSE    & sentences level   & articles level  \\
        \hline
     vagueness & $0.048$ & $0.012$  
     \\
     opinion  & $0.036$ & $0.010$ 
     \\ 
     detail         & $0.118$ & $0.046$ 
     \\
    \end{tabular}
    }
\end{center}
\caption{\label{tab:rmsecomp} Root Mean Square Error between \vago\ scores and \texttt{VAGO-N} scores, at sentence level ($N$=2,445) and at article level ($N$=100).}
\end{table}


Regarding our hypotheses, we computed correlations between the three \vago\ scores and the mean scores for the labels ``Subjective'', ``Opinions'', ``Exaggeration'', ``Fake News'' and ``False Information'', using both \vago\ and \vagon. Here we report the \vagon\ case only, as both versions give very similar results. As shown in Table \ref{tab:correlations}, we found positive correlations between scores of vagueness and opinion and the labels ``Subjective'', ``Opinions'', and ``Exaggeration'', but not for ``Fake News'' and ``False Information''. For all cases, however, we found a negative correlation between the score of detail to vagueness and the labels. The correlations are weak to moderate, but in the order of magnitude found in previous studies, and even higher in the labels ``Opinions'' and ``Subjective'' directly connected to \vago's opinion score. These results confirm that there is a stronger association between \vago\ markers of opinion and assessments of texts as ``Subjective'' than between those scores and assessments as ``Fake News'', a distinction not visible in \cite{Guelorget2021combining}'s analysis of the ISOT-False corpus, in which ``Biased'' was an annotation used indistinctly to refer to both ``Fake News'' and ``Opinions'' pieces that may not be fake.

\begin{table}[h]
{\small
\begin{tabular}[t]{l|c|c|c}
            &  vague  & opinion & detail  \\
\hline
Subjective                 &   $0.294^{**}$  & $0.339^{***}$  & $-0.380^{***}$ \\
\hline
 Opinions                  &  $0.266^{**}$ & $0.342^{***}$ & $-0.358^{***}$ \\
 \hline
 Exaggeration                 &  $0.217^{*}$ \phantom{.}   & $0.300^{**}$ \phantom{.} & $-0.232^{*}$ \phantom{..}  \\
\hline
 Fake News                  &  $0.134$ \phantom{...}   & $0.201$ \phantom{....}  & $-0.261^{**}$ \phantom{.} \\
\hline
 False Information                 &  $0.080$ \phantom{...}  & $0.139$ \phantom{....}  & $-0.303^{**}$ \phantom{.} \\

    \end{tabular}
}
\caption{Pearson correlations between labels' mean scores and \vagon\ scores ($^{*}$, $^{**}$, and $^{***}$ indicate $p$-value $<.05, <.01, < 0.001$).}\label{tab:correlations}
\end{table}

\section{Conclusion}

\pe{With only 100 documents, the corpus presented here is limited to train a classifier, but it is valuable in virtue of its rich set of annotations, and it can be used for further regression analyses concerning the ascription of the label ``Fake News'' relative to other labels. The analyses confirm that linguistic markers of subjectivity explain part of the variance in the ascription of labels such as ``Subjective'', ``Opinion'', ``Exaggeration'', but also ``Fake News''. Some labels in our study turns out to be uninformative (``Facts'', ``Places''), while others could be included (``Satirical'', to check for presence of humor). We refer the readers to the follow-up study \cite{faye2024exposing}, in which an adjusted set of 11 labels is used to analyze propaganda press.}

\clearpage

\section*{Limitations}

\pe{
All annotators have a higher-education degree (5 years or more after graduation), not necessarily representative of a larger and more diverse population.} 

\section*{Acknowledgements}

We thank three anonymous reviewers for helpful comments and feedback, and \textcolor{black}{Guillaume Gravier for his support}. This work was supported by the programs HYBRINFOX (ANR-21-ASIA-0003), FRONTCOG (ANR-17-EURE-0017), THEMIS (n°DOS0222794/00 and n°DOS0222795/00) and {PLEXUS (Marie Sk\l odowska-Curie Action, Horizon Europe Research and Innovation Programme, grant n°101086295). PE thanks Monash University for hosting him during the writing of this paper.

\section*{Declaration of contribution}

\textcolor{black}{All the authors contributed to the design, analysis, and discussion of the results. BI, GF, MC, and PE wrote the paper, which all authors read and revised together. Correspondence: benjamin.icard@lip6.fr, paul.egre@ens.psl.eu.}

\section*{References}\label{sec:reference}

\bibliographystyle{lrec-coling2024-natbib}
\bibliography{lrec-coling2024-example}

\end{document}